%% file: main.tex
\documentclass[a4paper,twoside]{article}

\usepackage{epsfig}
\usepackage{subcaption}
\usepackage{calc}
\usepackage{amssymb}
\usepackage{amstext}
\usepackage{amsmath}
\usepackage{amsthm}
\usepackage{multicol}
\usepackage{pslatex}
\usepackage{apalike}
\usepackage{booktabs}
\usepackage{hyperref}
\usepackage{xurl}
\usepackage[bottom]{footmisc}
\usepackage{SCITEPRESS}     

\begin{document}

\title{Debiasing Sentence Embedders through Contrastive Word Pairs}

 \author{\authorname{Philip Kenneweg\sup{1}, Sarah Schröder\sup{1}, Alexander Schulz\sup{1} and Barbara Hammer\sup{1}}
 \affiliation{\sup{1}CITEC, University of Bielefeld, Inspiration 1, 33615 Bielefeld, Germany}
 \email{\{pkenneweg, saschroeder, aschulz, bhammer\}@techfak.uni-bielefeld.de}}


\abstract{Over the last years, various sentence embedders have been an integral part in the success of current machine learning approaches to Natural Language Processing (NLP). Unfortunately, multiple sources have shown that the bias, inherent in the datasets upon which these embedding methods are trained, is learned by them. A variety of different approaches to remove biases in embeddings exists in the literature. Most of these approaches are applicable to word embeddings and in fewer cases to sentence embeddings. It is problematic that most debiasing approaches are directly transferred from word embeddings, therefore these approaches fail to take into account the nonlinear nature of sentence embedders and the embeddings they produce. It has been shown in literature that bias information is still present if sentence embeddings are debiased using such methods.
In this contribution, we explore an approach to remove linear and nonlinear bias information for NLP solutions, without impacting downstream performance. We compare our approach to common debiasing methods on classical bias metrics and on bias metrics which take nonlinear information into account.}

\onecolumn \maketitle \normalsize \setcounter{footnote}{0} \vfill

\section{\uppercase{Introduction}}
\label{sec:intro}
\input{intro}

\section{\uppercase{Related Work}}
\label{sec:related_work}
\input{related_work}

\section{\uppercase{Proposed Debiasing Approach}}
\label{sec:req}

\input{approach}

\section{\uppercase{Experimental Approach}}
\label{sec:experiments}

\input{experiments}

\section{\uppercase{Conclusions}}
\label{sec:conclusion}

\input{conclusion}

\textit{\section*{\uppercase{Acknowledgements}}}
We gratefully acknowledge funding by the BMWi (01MK20007E) in the project AI-marketplace.

\bibliographystyle{apalike}
{\small
\bibliography{references}}


\end{document}

%% file: intro.tex
In the last couple of years, the transformer architecture pioneered by \cite{DBLP:journals/corr/VaswaniSPUJGKP17} has enabled large pre-trained neural networks to efficiently tackle previously difficult NLP tasks with relatively few training examples. A common tool deployed to facilitate fast transfer of knowledge are Sentence Embedders which produce a vectorized representation of a given text input, where the original text $T$ can be of arbitrary length. However, longer text passages produce more diluted embeddings.

Sentence Embedders produce a summarization of the content that is easy to process for a multitude of different tasks by e.g.\ shallow neural networks. 
Many common architectures need a fine-tuning step on a specific task to achieve good performance \cite{DBLP:journals/corr/abs-1810-04805,DBLP:journals/corr/abs-1907-11692}.
Recent literature has shown that language models are inherently biased with regard to different protected attributes \cite{NIPS2016_a486cd07,DBLP:journals/corr/IslamBN16}. Commonly, investigated bias attributes are religion, gender, etc.\ \cite{may2019measuring}. 
Though the most common source of bias is the training data, other factors can have a mitigating or magnifying effect.

Many approaches to reduce bias in word/sentence embeddings exist \cite{DBLP:journals/corr/abs-2103-06413,DBLP:journals/corr/abs-2004-07667,DBLP:journals/corr/abs-1904-04047}, though most of these approaches were designed for word embeddings which produce linearly combinable embeddings. Sentence embeddings are based upon the vastly more complex transformer architecture, furthermore they are often processed by nonlinear multilayer networks. Many of the most common debiasing methods and metrics to evaluate bias in word/sentence embeddings are based upon the assumption of linearity in the embedding space. As this can no longer be guaranteed in sentence embeddings, new debiasing methods and ways to measure bias have to be considered.

In this paper we propose a new additional training objective that can debias sentence embeddings provided with only a few contrastive words that implicitly define the bias direction. Contrastive objectives, which are the basis for popular recent models such as CLIP \cite{radford2021learning} and the thereupon based DALL-E 2 \cite{ramesh2022hierarchical}, have proven to be particularly promising. 
Our proposed debiasing objective can be applied during the standard fine-tuning procedure required for many tasks or during the pre-training and provides better results than other debiasing procedures as we demonstrate, especially in the case of nonlinear bias.


%% file: related_work.tex
The most commonly used architecture for sentence embedders is the transformer architecture, which is pre-trained on large textual datasets.
Common objectives for transformer based pre-training are masked language modeling (MLM) and next sentence prediction (NSP) 
\cite{DBLP:journals/corr/abs-1810-04805}. 

A variety of previous works have shown that language models capture biases in their training data, which can manifest in text representations produced by these models \cite{NIPS2016_a486cd07,DBLP:journals/corr/IslamBN16}. This issue can further lead to issues in downstream tasks. For instance, Abid, Abubakar et al. \cite{gtp3-muslim-bias} showed that GPT-3 produces texts inheriting muslim-violence biases (e.g. in prompt completion). 
Such findings motivated many works to develop measures for biases in language models such as word and sentence embeddings \cite{NIPS2016_a486cd07,DBLP:journals/corr/IslamBN16} 
and debiasing algorithms \cite{NIPS2016_a486cd07,liang2020debiasing}. 

\subsection{Measuring bias in sentence embeddings}
Multiple different approaches to measuring bias in sentence embeddings exist. Some focus on the geometric relations of words in the embedding space \cite{DBLP:journals/corr/IslamBN16,may2019measuring,NIPS2016_a486cd07,DBLP:journals/corr/abs-1904-04047}, others on the influence of bias on classification, clustering or other downstream tasks \cite{gonen2019lipstick,DBLP:journals/corr/abs-1804-06876}.
In the course of our work, we use the classification and clustering test by \cite{gonen2019lipstick}. No universally agreed upon test to determine bias exists, and many tests used for bias measurement in sentence embeddings are adapted from word embeddings. This is accomplished by inserting words which are defining the bias space into carefully chosen neutral sentences. Since the sentence embedding space is in many cases highly nonlinear and more complex in contrast to many classical word embedding counterparts (e.g.\ word2vec \cite{mikolov2013efficient} uses a single layer without an activation function to produce the word embedding), it is not clear if the most common bias measuring methods applied in the word case can be applied to the sentence embedding case. Further, even in the word embedding context there exists criticism towards many bias metrics \cite{gonen2019lipstick,biasmeasures}.

\cite{gonen2019lipstick} propose a classification test for bias, where a classifier is trained to discriminate theoretically neutral words by stereotypical associations (in their case with gender). If the classifier can generalize these associations onto unseen embeddings, they are considered biased. The authors use a RBF-kernel SVM for classification.
The test can be easily expanded to sentence embedding by inserting such words into neutral sentences and then classifying on these sentences. In our experiments, we use a list of occupations used in the work of \cite{NIPS2016_a486cd07} as theoretically neutral words compared to gender attributes.
Furthermore, the choice of classifier influences which kinds of biases are detected, linear classifiers can only detect linear biases, whereas nonlinear classifiers can be used to detect more complex biases. The classification test can only detect the presence of bias and the relative amount, but it can not guarantee that no bias is present. 

\subsection{Removing bias in sentence embeddings}
Word embeddings are well researched in comparison to sentence embeddings with regards to their bias. Since most word embeddings methods are inherently linear, the approaches used for debiasing cannot be directly applied to sentence embeddings.\newline
Most recent work on removing bias in sentence embeddings is focused on removing the bias while treating the neural network as a black box and only applying their debiasing procedure post-hoc on the sentence embeddings \cite{liang2020debiasing,DBLP:journals/corr/abs-2103-06413}.
Other work tries to debias sentence embedders by retraining them on unbiased data \cite{zhao2019gender}. Obtaining large quantities of unbiased data however proves difficult. 

In this paper, we will directly retrain the network using a custom loss function, assuming that the capabilities of transformers to understand complex relations also make them perform well at debiasing. \newline
\cite{liang2020debiasing} propose the Sent-Debias approach. It utilizes 
PCA to capture the gender dimension in a large variety of sentences, by replacing gender sensitive words by their counterparts and computing the difference of the produced sentence embedding. Furthermore, many different naturally occurring sentences from a text corpus are utilized, thus capturing more complexity of sentence embeddings than by using purely simple sentence templates.
\newline
Null-It-Out \cite{DBLP:journals/corr/abs-2004-07667} approaches the problem differently by looking at an SVM classifier and a corresponding bias related task. Then their so called iterative nullspace projection algorithm is performed that results in the SVM classifier no longer producing meaningful predictions.They also highlight that the proposed approach only works for linear classifiers and the corresponding information can be easily recovered by a nonlinear model. We will compare to this model in our experiments.
\newline
FairFil \cite{DBLP:journals/corr/abs-2103-06413} has a similar approach to generating contrastive sentences as our approach, but uses only a comparatively small set of manually selected sentences. Moreover, the neural network is treated as a black box on which the generated sentence embeddings are debiased using an extra filter. 

%% file: approach.tex
Approaches like Sent-Debias \cite{liang2020debiasing} or Null It Out \cite{DBLP:journals/corr/abs-2004-07667} are only capable of removing linear biases and dependencies in the sentence embeddings.
Simple three layer deep neural networks are able to recover most bias from examples that are debiased using these methods, see Tables \ref{Fig:colaocc}, \ref{Fig:sst2occ} and \ref{Fig:qnliocc}.
For this reason we try to remove linear and nonlinear information present in the embeddings regarding the bias, while retaining performance on a variety of classification tasks.
In literature \cite{liang2020debiasing} the importance of using a wide variety of different sentences in a semi-supervised
fashion is highlighted to generate better debiasing. We will follow this concept in the present paper.
In contrast to \cite{liang2020debiasing} using linear projections as calculated by a PCA to perform their debiasing, we will focus in this work on potential improvements by training the whole network with an additional cost function during fine tuning, pre-training or both.

\subsection{Definition}
First we choose word pairs $K_1, K_2, .. K_n$ that contrastively define the bias subspace. For example:\\ $K_1 = $ [men, women], $K_2 = $ [boy, girl], $K_3 = $ [muslim, christian, jew]. \newline
The words chosen should only differ in their meaning by the targeted bias. \newline
In the next step a large text data set $D$
is searched for occurrences of any of these words $k\in K_i \exists i$. Whenever a sentence $S$ is found in which one of these words $k$ occurs, the original sentence $S_o$ and a variation $S_c$, where the word $k\in K_i$ is replaced by one of its counterparts in $K_i$, is added as a pair $S_o,S_c$ to the debias training examples. If multiple counterparts  $k\in K_i$ are available as is the case in our religion example one is selected at random. 
Accordingly, we propose the following loss objective upon which the network is additionally trained:
\begin{equation}
    L(S_o,S_c) = \|E(S_o) - E(S_c)\| 
    \label{eq:loss}
\end{equation}
And the overall debiasing loss for one training epoch is:
\begin{equation}
    L = \sum_{S_o, S_c}^{} L(S_o,S_c)
    \label{eq:losssum}
\end{equation}
Thereby, $E()$ is the embedding function produced by the network and $\|.\|$ is the euclidean norm. The intuition behind this objective is to penalize the network for producing different/biased embeddings if only the gender information differs in the sentences. Overall this incentivizes the network to not convey any bias related information in its embedding.
\newline
The additional objective function we propose can be applied during fine-tuning on a final classification task, or during pre-training of the model.

It is always necessary to perform another training objective (for example the fine tuning task, or a variety of different pre-training task) concurrently since a clear solution to minimize the loss described in equation \ref{eq:losssum} is to produce the same embedding/shrink all embeddings for every input. 
\newline Our additional objective is semi-supervised to leverage the capability of understanding complex relations of transformers with large amounts of data. 
All in all, we propose three schemes for debiasing by augmenting the training with equation \ref{eq:losssum}:
\begin{itemize}
    \item include the proposed cost function during pre-training, further referred to as $pre^p$
    \item include the proposed cost function during fine-tuning, further referred to as $fine^p$
    \item include the proposed cost function during pre-training and fine-tuning, further referred to as $prefine^p$
\end{itemize}
We utilize the letter $p$ to refer to our approach \emph{pairwise contrastive bias reduction}.

%% file: experiments.tex
In this section we detail our employed experimental design to investigate the effects of our proposed approaches $fine^p$, $pre^p$ and $prefine^p$ with respect to the reduction of bias and the performance on downstream tasks.
We utilize the BertHugginface library \cite{wolf2020huggingfaces} for implementation and the pre-trained Bert model ('bert-base-uncased') for all experiments in order to reduce training time.

\subsection{Generating Contrastive Sentences} \label{sec:gensent}
In order to produce comparable results, we follow a large line of work of in the literature and employ gender attributes for debiasing evaluation. The word pairs we use to define the gender dimensions could for example be:\newline women - men and girl - boy. \newline A larger selection of word pairs consisting of 11 pairs is used for our experiments. 
For each bias attribute a large amount of sentences (20,000 in our case) of the News Corpus Multi-News \cite{fabbri2019multinews} was found in which the bias definition words are present. These are then utilized as sentences $S_o$ and $S_c$ for our proposed approaches.

\subsection{Datasets}

The Glue dataset by \cite{wang2019glue} is a collection of other datasets and is widely used to evaluate common natural language processing capabilites of a variety of networks. All datasets used are the version provided by tensorflow-datasets 4.0.1.




\subsection{Implementation Details} \label{sec:impl}

\emph{Occupation Task:}\cite{gonen2019lipstick} propose a classification test to determine bias, where a classifier $C$ is trained to discriminate for embeddings of occupations $E(w)$, whether $E(w)$ is typically male or female. 
A high accuracy in this setting corresponds to a stereotype present in the embedding. 


This test can be directly expanded to the sentence embedding context by inserting such words into neutral sentences  and then classifying these sentences. In our experiments, we utilize a list of occupations used in the work of \cite{NIPS2016_a486cd07} as theoretically neutral words compared to gender attributes. Since these occupations are rated by \cite{NIPS2016_a486cd07} by how stereotypical they are male/female, the classification task from \cite{NIPS2016_a486cd07} is modified to be a regression to these ratings.

This test is highly relevant for the purposes of this paper since it can easily be modified to be only able to distinguish linear bias information (by using only a single layer without an activation function), as well as nonlinear information (using an MLP). Furthermore the reported results of this test have low standard deviations compared to the SEAT \cite{liang2020debiasing} test, additionally the results of the SEAT test have low statistical significance as reported in \cite{gonen2019lipstick}.

In our work, we implement the occupation task by the BERT architecture sentence embedder, which was pre-trained and/or fine-tuned using the parameters supplied by the BERT paper. During the training of the bias regressor this part of the network is kept frozen. The produced embeddings are then fed into a Multi Layer Perceptron. In the linear case the MLP consists of just a single neuron with a sigmoid activation function. In the nonlinear case it consists of 3 Dense Layers with 20 neurons each and Rectified Linear Unit (RELU) activation functions inbetween. Again a single neuron with a sigmoid activation function acts as the output.

Each model is trained for 50 epochs on the training data or until the validation accuracy does not improve for 5 epochs. The optimizer used is SGD with a learning rate of 0.01. The utilized loss function to compute the regression loss is the Mean Squared Error (MSE).

Finally, the resulting loss on the test set after training is the score used in Figure \ref{fig:biassteps}, \ref{fig:biassteps2} and Table \ref{Fig:colaocc}, \ref{Fig:sst2occ}, \ref{Fig:qnliocc}.


\subsubsection{Pre training.}
\begin{figure}[t]
	\centering
	\includegraphics[scale=0.45]{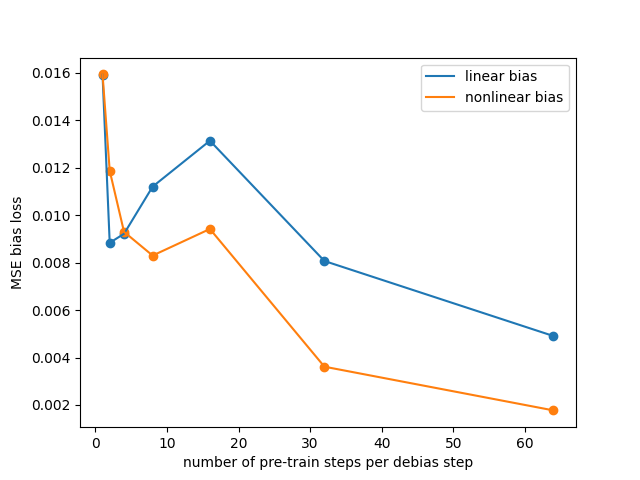}
	\caption{Number of pre-training steps per debiasing step vs detected bias. Higher Occupation task loss denotes less bias present in the embedding.}
	\label{fig:biassteps}
\end{figure}

During additional pre-training of $pre^p$ and $prefine^p$ we use the pre-trained Bert Model and train it on a news corpus using the MLM task concurrently with our debiasing method for gender for 400,000 training steps. Using an already pre-trained model is done to speed up convergence. \newline 
If the debias loss is balanced with the MLM loss correctly, only the bias relevant information should be removed from the model, while retaining high performance on the MLM task and possible downstream tasks.
In order to estimate a good balance between these two objectives, we evaluate different ratios of MLM steps per debias step on the Occupation task (see Figure \ref{fig:biassteps}).
The debiasing performance clearly drops when the number of pre-training steps per debias step increases. However, the MLM pre-training loss is not affected(not depicted). Accordingly, we use the ratio of 1:1 for the training of $pre^p$ and $prefine^p$. 

\subsubsection{Fine-tuning.}
During fine-tuning of $prefine^p$ and $fine^p$ we employ the pre-trained Bert Model provided by Huggingface \cite{wolf2020huggingfaces} or $pre^p$.

\begin{figure}[ht]
	\centering

	\includegraphics[scale=0.45]{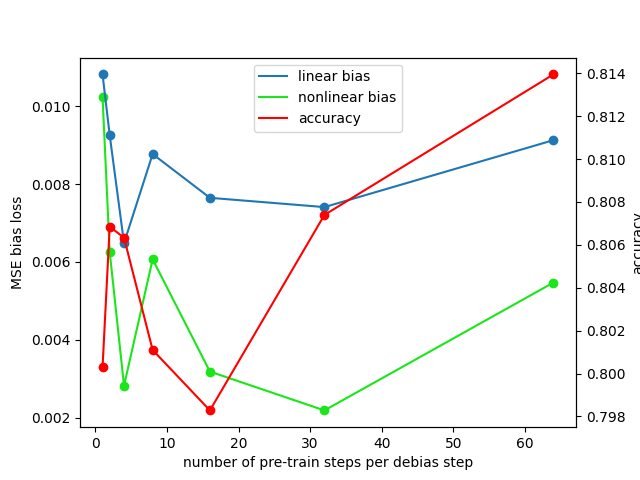}
	\caption{Number of fine-tune steps per debiasing step vs detected bias. Higher bias classifier loss denotes less bias present in the embedding. The red line denotes the accuracy on the downstream task. The blue line shows the bias loss of the linear model and the yellow line denotes the bias loss of the nonlinear model. The bias scores and accuracies are averaged over 5 runs.}
	\label{fig:biassteps2}
\end{figure}

In order to estimate the same balance between our debiasing objective and the fine-tuning objective, we perform a search on the relative number of steps on the glue task vs steps on the debiasing task. Since we are in a fine-tuning scenario, we can now incorporate the accuracy on a downstream task for our search. The result is shown in Figure \ref{fig:biassteps2}.
A clear downward trend can be observed for the nonlinear bias score (lower score equals more bias present in the embedding) and less clearly for the linear one, in case of more fine-tuning steps per debiasing step. While the accuracy does seem to be negatively affected in this case, this effect is rather minor (in the range of 1-2\%). Following this, the same amount of steps with our custom training objective are performed as on the classification task during fine-tuning (1:1).


For each fine-tuning task, the pre-trained BERT architecture is used and a single new fully connected layer after the sentence embeddings is added, then the whole network is trained on the fine-tuning task. All fine-tuning runs are trained for 5 epochs with 7,200 training examples per epoch. Longer training times and hyperparameter tuning could have resulted in better performance, but since the topic of this paper is debiasing these models, only the relative performance of the different training setups matter. The used hyper-parameters are the same as recommended in the original BERT paper \cite{DBLP:journals/corr/abs-1810-04805}.
All fine-tuning experiments are performed 5 times and their averages are reported. This is done to reduce variation in results, since BERT fine-tuning is very sensitive to small changes. The same run can vary in performance up to 5\% just due to non-deterministic training on the graphics card. 

\subsection{Experimental Setup}

In the following, we evaluate the performance of our debiasing approach and investigate possible performance drops on downstream tasks

For the glue tasks, we consider a pre-trained Bert model which we fine-tune on one downstream task together with different debiasing approaches: pre-trained and fine-tuned Bert without debiasing (referred to as \emph{original Bert}), pre-trained and fine-tuned Bert with subsequent debiasing by the Sent-Debias approach (referred to as \emph{Sent-Debias}), pre-trained and fine-tuned Bert with subsequent debiasing by Null-it-Out (referred to as \emph{Null-it-Out}), pre-trained Bert together with our proposed $fine^p$ debiasing (which involves fine-tuning; referred to as $fine^p$) and pre-trained Bert together with our proposed $prefine^p$ debiasing (which involves fine-tuning; referred to as $prefine^p$). Each of these combinations is, after performing the debiasing (see the descriptions in sec.\ \ref{sec:impl}), evaluated on a test set of the according downstream task giving the \emph{accuracy} values shown in the tables. 
Subsequently, we employ the occupation task in order to estimate the amount of bias present in the models. This is done by adding (a) new layer(s), which parameters are trained to estimate the gender of a job embedded in a sentence (see secs.\ \ref{sec:gensent} and \ref{sec:impl} for more details). According to the number of these new layers we refer to \emph{linear bias} (one layer) or \emph{nonlinear bias} (multiple layers, with activation functions in between) and high values imply low bias.
The results of these experiments are depicted in section \ref{sec:glue}.

\begin{table*}[t]
  \centering
  \caption{Classification accuracies, linear and nonlinear bias (loss on the occupation task) on the Corpus of Linguistic Acceptability task after fine tuning. All fine tuning runs were performed five times and scores averaged. Higher bias scores indicate lower bias present in the embeddings. Baseline bias score (average prediction) equals to $0.0197$.}
  \label{Fig:colaocc}
  \begin{tabular}{c ccc}
    \toprule
method & accuracy $\uparrow$ & linear bias $\uparrow$ & nonlinear bias $\uparrow$ \\
 \cmidrule(r){1-1} \cmidrule(l){2-4}
original Bert &  \textbf{0.783} $\pm$ 0.0072 &  0.00677 $\pm$ 0.00106  & 0.00315 $\pm$ 0.00093 \\
Sent-Debias & 0.781 $\pm$ 0.0095 &  0.00705 $\pm$ 0.00085  & 0.00258 $\pm$ 0.00085 \\
Null-It-Out & 0.765 $\pm$ 0.0154 & \textbf{0.01635} $\pm$ 0.00116 & 0.00593 $\pm$ 0.00524 \\\
$fine^p$ & 0.765 $\pm$ 0.0084 & 0.00806 $\pm$ 0.00375 & 0.00393 $\pm$ 0.00386 \\
$prefine^p$ & 0.771 $\pm$ 0.0147 & 0.01100 $\pm$ 0.00452 &  \textbf{0.01310} $\pm$ 0.00617\\
    \bottomrule
  \end{tabular}
\end{table*}

In a second series of experiments we investigate debiasing capabilities on pre-trained models that are not fine-tuned. These are displayed in section \ref{sec:pretrModels}.



\section{Results}

\begin{table*}[t]
  \centering
  \caption{Classification accuracies, linear and nonlinear bias on the Stanford Sentiment Treebank task after fine tuning. All fine tuning runs were performed five times and their scores averaged. Baseline bias score (average prediction) equals $0.0197$.}
  \label{Fig:sst2occ}
  \begin{tabular}{c ccc}
    \toprule
method & accuracy $\uparrow$ & linear bias $\uparrow$ & nonlinear bias $\uparrow$  \\
 \cmidrule(r){1-1} \cmidrule(l){2-4}
original Bert &  \textbf{0.870} $\pm$ 0.0092  &  0.00473 $\pm$ 0.00075  & 0.00259 $\pm$ 0.00044 \\
Sent-Debias &  0.851  $\pm$ 0.0084 &  0.00482 $\pm$ 0.00072 & 0.00254 $\pm$ 0.00047 \\
Null-It-Out & 0.864 $\pm$ 0.0247 & \textbf{0.01539} $\pm$ 0.00105  & 0.00110 $\pm$ 0.00055  \\
$fine^p$ & 0.858 $\pm$ 0.0122  & 0.00553 $\pm$ 0.00269 & 0.00288  $\pm$ 0.00187\\
$prefine^p$ & 0.851 $\pm$ 0.0434  & 0.00989 $\pm$ 0.00101 &  \textbf{0.00856} $\pm$ 0.00236 \\
    \bottomrule
  \end{tabular}
\end{table*}
This section displays the results of the four experimental series described previously.

\subsection{GLUE} \label{sec:glue}
For evaluation, 3 Glue \cite{wang2019glue} tasks (CoLA, SST2, QNLI) are considered. The results are described in the following three subsections. 


\subsubsection{Corpus of Linguistic Acceptability}

The results for the CoLA data set are summarized in Table \ref{Fig:colaocc}.
The accuracy for the CoLA task does not change strongly when fine-tuning or pre-training with our additional debiasing objective. The best performing model is the original BERT. While Sent-Debias is able to increase performance on the linear occupation task (i.e.\ reduce bias), it actually performes worse than the original BERT on the nonlinear occupation task. Null-It-Out does significantly decrease the linear bias detected and even decreases the nonlinear bias found. Both proposed debiasing approaches are able to decrease the evaluated bias significantly, compared to the original BERT model. Especially the $prefine$ method achieves significantly higher debias scores in the nonlinear domain, whereas the $fine$ method only decreases the measured bias slightly.

\subsubsection{Stanford Sentiment Treebank}

The results for the SST2 data set are display in Table \ref{Fig:sst2occ}.
The best accuracy on this task is achieved by the original BERT. $prefine^p$ performes $1.9\%$ worse on accuracy while achieving almost $100\%$ better performance on the linear debias score and more than $100\%$ improvement on the nonlinear debias score. Interestingly, $prefine^p$ performs better on the occupation task than $fine^p$, while Sent-Debias along with the original BERT perform worse than $prefine^p$.
Null-It-Out performs the best on the linear bias score, but decreases performance on the nonlinear bias score.

\subsubsection{Stanford Question Answering Dataset}
The results for the QNLI data set are given in Table \ref{Fig:qnliocc}.
Here, the best accuracy is achieved by the original BERT. The best debiasing performance is achieved by $prefine^p$ in linear and nonlinear bias. $fine^p$ and Sent-Debias performed very similar to the original BERT. While Null-It-Out performs well on the linear bias score, it fails to debias the occupations nonlinearly.

\begin{table*}[t]
  \centering
  \caption{Classification accuracies, linear and nonlinear bias on the Stanford Question Answering Dataset task after fine tuning. Baseline bias score (average prediction) equals $0.0197$.}
  \label{Fig:qnliocc}
  \begin{tabular}{c ccc}
    \toprule
method & accuracy $\uparrow$ & linear bias $\uparrow$ & nonlinear bias $\uparrow$ \\
 \cmidrule(r){1-1} \cmidrule(l){2-4}
original Bert &  \textbf{0.817} $\pm$ 0.0060 &  0.0103 $\pm$  0.0023  & 0.0070 $\pm$ 0.0041 \\
Sent-Debias & 0.803 $\pm$ 0.0056  &  0.0108 $\pm$ 0.0025 & 0.0049 $\pm$ 0.0024  \\
Null-It-Out & 0.814 $\pm$ 0.0073 & 0.0166 $\pm$ 0.0015 & 0.0029 $\pm$ 0.0047 \\
$fine^p$ & 0.807 $\pm$ 0.0153  & 0.0089 $\pm$ 0.0032 & 0.0053 $\pm$ 0.0032 \\
$prefine^p$ & 0.807 $\pm$ 0.0100  & \textbf{0.0202} $\pm$ 0.0020 &  \textbf{0.0191} $\pm$ 0.0009 \\
    \bottomrule
  \end{tabular}
\end{table*}

\subsection{Pre-trained Models} \label{sec:pretrModels}
A variety of machine learning models try to use sentence embeddings without any fine-tuning. 
Hence, we investigate the performance of various debiasing methods without any fine-tuning to a specific task.

\begin{table}
  \centering
  \caption{linear and nonlinear bias of the pretrained embedding models without any finetuning to a specific task. Higher bias scores indicate lower bias present in the embeddings. Baseline indicates the average prediction.}
  \label{Fig:pretrain}
  \begin{tabular}{c cc}
    \toprule
method  & linear bias $\uparrow$ & nonlinear bias $\uparrow$ \\
 \cmidrule(r){1-1} \cmidrule(l){2-3}
original Bert  &  0.0047 &  0.0018 \\
Sent-Debias  &  0.0046 &  0.0020  \\
Null-It-Out & 0.0157 & 0.0021 \\
$pre^p$  & \textbf{0.0159} & \textbf{0.0159}  \\
baseline & 0.0197 & 0.0197 \\
    \bottomrule
  \end{tabular}
\end{table}

In Table \ref{Fig:pretrain}, $pre^p$  performs the best with regards to the bias score. As no task is associated with pure sentence embeddings no accuracy can be given to compare the performance of the embeddings. For performance comparison see Tables \ref{Fig:colaocc}, \ref{Fig:sst2occ} and \ref{Fig:qnliocc}, here $pre^p$ is further trained with the additional debias loss to produce $prefine^p$. It can be seen, that the performance of $prefine^p$ is lower than the original BERT performance on the Glue tasks by 1.3\% on average.

\subsubsection{Summary}

Overall, the additional debiasing objectives (our proposed ones and sent-debias) do not significantly impact the glue tasks performance compared to the orignal BERT model. Models where our debiasing objective was applied during fine-tuning or fine-tuning and pre-training report only minor accuracy losses. 

In summary, the best performing version of our debiasing algorithms is $prefine^p$. It beats the original BERT and Sent-Debias in all cases regarding bias score on the occupation task, while the accuracy on the glue tasks is mostly retained. 
While Null-It-Out performs well on the linear debiasing measure, it completely fails to reduce the nonlinear measure.
$fine^p$ often improves the bias score (by a smaller margin than $prefine^p$) and is a valid option due to being simpler and computationally cheaper than $prefine^p$.

To further gain an intuition into the proposed debiasing method and its effects on the embedding space, we visualize the average sentence embeddings of the occupations from the bias classification task using DeepView \cite{deepview}. 
In the linear case it can be seen that the lighter areas, denoting more uncertainty on the part of the classifier, have grown compared to the original example and that the distinction between the two classes is not as large for the debiased case.
Interestingly the performance on the classification task of the nonlinear classifier has decreased significantly after debiasing from 99.84 \% to 83.75 \% accuracy (The average classifier which always produces as an output the male class achieves an accuracy of 83.33), even though the visualization does not seem to have changed extensively.

\begin{figure}[t]
	\centering
	\includegraphics[scale=0.18]{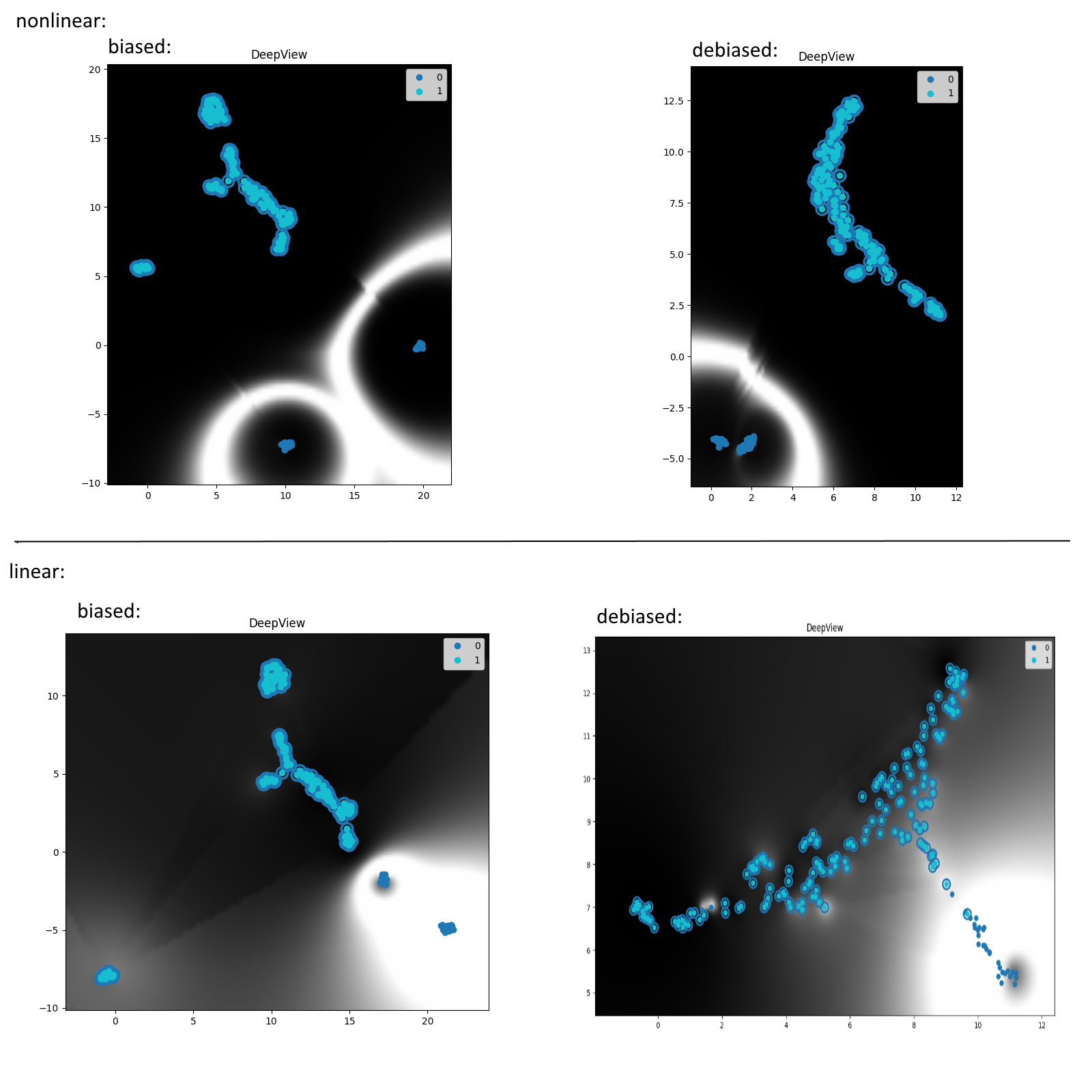}
    \caption{DeepView visualization of the original BERT embedding vs $pre^p$ embeddings. Dark areas denote regions of space where the predictor has a high confidence value, whereas lighter areas indicate greater uncertainty. Light blue points indicate female occupations and dark blue points indicate male occupations}
	\label{fig:deepview}
\end{figure}

%% file: conclusion.tex
In this paper we presented an easy to implement additional training objective that can be applied during pre-training and/or fine-tuning of the network. It decreases the bias measured by the occupation task clearly, while not impacting the accuracy on downstream task. Furthermore we show that, using this method, we can strongly reduce nonlinear gender information in contrast to most other debiasing algorithms, which can otherwise easily be recovered by a multilayer perceptron. 

Overall, further research on this topic is required, as there is still no consensus regarding suitable metrics for detecting and comparing biases in sentence embeddings. Even though our debiasing approach seems promising in erasing bias related information in sentence embeddings, it is not able to reliably erase all information present. Thus debiasing algorithms which are able to completely, precisely and reliably erase linear and nonlinear information targeting a certain bias concept are still needed.

All code for this paper can be found on \url{https://github.com/TheMody/Debiasing-Sentence-Embedders-through-contrastive-word-pairs}.

%% file: main.bbl
\begin{thebibliography}{}

\bibitem[Abid et~al., 2021]{gtp3-muslim-bias}
Abid, A., Farooqi, M., and Zou, J. (2021).
\newblock Persistent anti-muslim bias in large language models.
\newblock In {\em Proceedings of the 2021 AAAI/ACM Conference on AI, Ethics,
  and Society}, AIES '21, page 298–306, New York, NY, USA. Association for
  Computing Machinery.

\bibitem[Bolukbasi et~al., 2016]{NIPS2016_a486cd07}
Bolukbasi, T., Chang, K.-W., Zou, J.~Y., Saligrama, V., and Kalai, A.~T.
  (2016).
\newblock Man is to computer programmer as woman is to homemaker? debiasing
  word embeddings.
\newblock In Lee, D., Sugiyama, M., Luxburg, U., Guyon, I., and Garnett, R.,
  editors, {\em Advances in Neural Information Processing Systems}, volume~29,
  pages 4349--4357. Curran Associates, Inc.

\bibitem[Caliskan et~al., 2017]{DBLP:journals/corr/IslamBN16}
Caliskan, A., Bryson, J.~J., and Narayanan, A. (2017).
\newblock Semantics derived automatically from language corpora contain
  human-like biases.
\newblock {\em Science}, 356(6334):183--186.

\bibitem[Cheng et~al., 2021]{DBLP:journals/corr/abs-2103-06413}
Cheng, P., Hao, W., Yuan, S., Si, S., and Carin, L. (2021).
\newblock Fairfil: Contrastive neural debiasing method for pretrained text
  encoders.
\newblock {\em CoRR}, abs/2103.06413.

\bibitem[Devlin et~al., 2018]{DBLP:journals/corr/abs-1810-04805}
Devlin, J., Chang, M., Lee, K., and Toutanova, K. (2018).
\newblock {BERT:} pre-training of deep bidirectional transformers for language
  understanding.
\newblock {\em Proceedings of the 2019 Conference of the North American Chapter
  of the Association for Computational Linguistics: Human Language
  Technologies}, abs/1810.04805.

\bibitem[Fabbri et~al., 2019]{fabbri2019multinews}
Fabbri, A.~R., Li, I., She, T., Li, S., and Radev, D.~R. (2019).
\newblock Multi-news: a large-scale multi-document summarization dataset and
  abstractive hierarchical model.

\bibitem[Gonen and Goldberg, 2019]{gonen2019lipstick}
Gonen, H. and Goldberg, Y. (2019).
\newblock Lipstick on a pig: {D}ebiasing methods cover up systematic gender
  biases in word embeddings but do not remove them.
\newblock In {\em Proceedings of the 2019 Conference of the North {A}merican
  Chapter of the Association for Computational Linguistics: Human Language
  Technologies}, pages 609--614, Minneapolis, Minnesota. Association for
  Computational Linguistics.

\bibitem[Liang et~al., 2020]{liang2020debiasing}
Liang, P.~P., Li, I.~M., Zheng, E., Lim, Y.~C., Salakhutdinov, R., and Morency,
  L.-P. (2020).
\newblock Towards debiasing sentence representations.

\bibitem[Liu et~al., 2019]{DBLP:journals/corr/abs-1907-11692}
Liu, Y., Ott, M., Goyal, N., Du, J., Joshi, M., Chen, D., Levy, O., Lewis, M.,
  Zettlemoyer, L., and Stoyanov, V. (2019).
\newblock Roberta: {A} robustly optimized {BERT} pretraining approach.
\newblock {\em CoRR}, abs/1907.11692.

\bibitem[Manzini et~al., 2019]{DBLP:journals/corr/abs-1904-04047}
Manzini, T., Yao~Chong, L., Black, A.~W., and Tsvetkov, Y. (2019).
\newblock Black is to criminal as caucasian is to police: Detecting and
  removing multiclass bias in word embeddings.
\newblock In {\em Proceedings of the 2019 Conference of the North {A}merican
  Chapter of the Association for Computational Linguistics: Human Language
  Technologies, Volume 1}, pages 615--621, Minneapolis, Minnesota. Association
  for Computational Linguistics.

\bibitem[May et~al., 2019]{may2019measuring}
May, C., Wang, A., Bordia, S., Bowman, S.~R., and Rudinger, R. (2019).
\newblock On measuring social biases in sentence encoders.

\bibitem[Mikolov et~al., 2013]{mikolov2013efficient}
Mikolov, T., Chen, K., Corrado, G., and Dean, J. (2013).
\newblock Efficient estimation of word representations in vector space.

\bibitem[Radford et~al., 2021]{radford2021learning}
Radford, A., Kim, J.~W., Hallacy, C., Ramesh, A., Goh, G., Agarwal, S., Sastry,
  G., Askell, A., Mishkin, P., Clark, J., et~al. (2021).
\newblock Learning transferable visual models from natural language
  supervision.
\newblock In {\em International Conference on Machine Learning}, pages
  8748--8763. PMLR.

\bibitem[Ramesh et~al., 2022]{ramesh2022hierarchical}
Ramesh, A., Dhariwal, P., Nichol, A., Chu, C., and Chen, M. (2022).
\newblock Hierarchical text-conditional image generation with clip latents.
\newblock {\em OpenAI papers}.

\bibitem[Ravfogel et~al., 2020]{DBLP:journals/corr/abs-2004-07667}
Ravfogel, S., Elazar, Y., Gonen, H., Twiton, M., and Goldberg, Y. (2020).
\newblock Null it out: Guarding protected attributes by iterative nullspace
  projection.
\newblock In {\em Proceedings of the 58th Annual Meeting of the Association for
  Computational Linguistics}, pages 7237--7256, Online. Association for
  Computational Linguistics.

\bibitem[Schröder et~al., 2023]{biasmeasures}
Schröder, S., Schulz, A., Kenneweg, P., and Hammer, B. (2023).
\newblock So can we use intrinsic bias measures or not?
\newblock In {\em International Conference on Pattern Recognition Applications
  and Methods}.

\bibitem[Schulz et~al., 2020]{deepview}
Schulz, A., Hinder, F., and Hammer, B. (2020).
\newblock Deepview: Visualizing classification boundaries of deep neural
  networks as scatter plots using discriminative dimensionality reduction.
\newblock {\em Proceedings of the Twenty-Ninth International Joint Conference
  on Artificial Intelligence}.

\bibitem[Vaswani et~al., 2017]{DBLP:journals/corr/VaswaniSPUJGKP17}
Vaswani, A., Shazeer, N., Parmar, N., Uszkoreit, J., Jones, L., Gomez, A.~N.,
  Kaiser, L.~u., and Polosukhin, I. (2017).
\newblock Attention is all you need.
\newblock In Guyon, I., Luxburg, U.~V., Bengio, S., Wallach, H., Fergus, R.,
  Vishwanathan, S., and Garnett, R., editors, {\em Advances in Neural
  Information Processing Systems}, volume~30. Curran Associates, Inc.

\bibitem[Wang et~al., 2019]{wang2019glue}
Wang, A., Singh, A., Michael, J., Hill, F., Levy, O., and Bowman, S.~R. (2019).
\newblock {GLUE}: A multi-task benchmark and analysis platform for natural
  language understanding.
\newblock In the Proceedings of ICLR.

\bibitem[Wolf et~al., 2020]{wolf2020huggingfaces}
Wolf, T., Debut, L., Sanh, V., Chaumond, J., Delangue, C., Moi, A., Cistac, P.,
  Rault, T., Louf, R., Funtowicz, M., Davison, J., Shleifer, S., von Platen,
  P., Ma, C., Jernite, Y., Plu, J., Xu, C., Scao, T.~L., Gugger, S., Drame, M.,
  Lhoest, Q., and Rush, A.~M. (2020).
\newblock Huggingface's transformers: State-of-the-art natural language
  processing.

\bibitem[Zhao et~al., 2019]{zhao2019gender}
Zhao, J., Wang, T., Yatskar, M., Cotterell, R., Ordonez, V., and Chang, K.-W.
  (2019).
\newblock Gender bias in contextualized word embeddings.

\bibitem[Zhao et~al., 2018]{DBLP:journals/corr/abs-1804-06876}
Zhao, J., Wang, T., Yatskar, M., Ordonez, V., and Chang, K.-W. (2018).
\newblock Gender bias in coreference resolution: Evaluation and debiasing
  methods.
\newblock In {\em Proceedings of the 2018 Conference of the North {A}merican
  Chapter of the Association for Computational Linguistics: Human Language
  Technologies, Volume 2}, pages 15--20, New Orleans, Louisiana. Association
  for Computational Linguistics.

\end{thebibliography}
